\pdfoutput=1

\documentclass[11pt]{article}
\RequirePackage[]{emnlp2021}

\RequirePackage{xcolor}

\RequirePackage[disable,textsize=footnotesize,color=yellow!10,linecolor=orange]{todonotes}
\RequirePackage{amsmath,amssymb,bm,stmaryrd}
\makeatletter
\g@addto@macro{\normalsize}{%
\setlength{\abovedisplayskip}{3pt plus1pt}%
\setlength{\abovedisplayshortskip}{3pt plus1pt}%
\setlength{\belowdisplayskip}{3pt plus1pt}%
\setlength{\belowdisplayshortskip}{3pt plus1pt}}
\let\c@table\c@figure
\makeatother

\usepackage{times}
\usepackage{latexsym}
\usepackage{graphicx}
\usepackage[T1]{fontenc}

\usepackage[utf8]{inputenc}
\usepackage{pgfplots}
\usepackage{bbold}
\usepackage{microtype}

\RequirePackage{adjustbox}

\DeclareMathOperator{\SO}{SO}
\DeclareMathOperator{\mBERT}{bert}
\DeclareMathOperator*{\argmax}{argmax}

\def\En{\ensuremath{\text{En}}}
\def\Fr{\ensuremath{\text{Fr}}}
\def\KG{\ensuremath{\text{KG}}}
\def\KbcOne{\textsc{KGC\-one}}
\def\KbcUnion{\textsc{KGC\-union}}
\def\shortname{\textsc{Align\-KGC}}

\title{Multilingual Knowledge Graph Completion
\\with Joint Relation and Entity Alignment}

\author{
Harkanwar Singh$^{1}$, 
{Prachi Jain}$^1$, 
Mausam$^1$ {\normalfont and} 
Soumen Chakrabarti$^2$
\\ 
$^1$ Indian Institute of Technology Delhi \\
$^2$ Indian Institute of Technology Bombay  \\
\{harkanwarsingh1841, p6.jain\}@gmail.com,
mausam@iitd.ac.in,
soumen@iitb.ac.in
}

\begin{document}
\maketitle

\begin{abstract}
Knowledge Graph Completion (KGC) predicts missing facts in an incomplete Knowledge Graph. Almost all of existing KGC research is applicable to only one KG at a time, and in one language only. However, different language speakers may maintain separate KGs in their language and no individual KG is expected to be complete.  Moreover, common entities or relations in these KGs have different surface forms and IDs, leading to ID proliferation. Entity alignment (EA) and relation alignment (RA) tasks resolve this by recognizing pairs of entity (relation) IDs in different KGs that represent the same entity (relation). This can further help prediction of missing facts, since knowledge from one KG is likely to benefit completion of another. High confidence predictions may also add valuable information for the alignment tasks.

In response, we study the novel task of jointly training multilingual KGC, relation alignment and entity alignment models. We present \shortname, which uses some seed alignments to jointly optimize all three of KGC, EA and RA losses. A key component of \shortname{} is an embedding-based soft notion of asymmetric overlap defined on the (subject, object) set signatures of relations -- this aids in better predicting relations that are equivalent to or implied by other relations.  Extensive experiments with DBPedia in five languages establish the benefits of joint training for all tasks, achieving 10-32 MRR improvements of \shortname{} over a strong state-of-the-art single-KGC system completion model over each monolingual KG . Further, \shortname{} achieves reasonable gains in EA and RA tasks over a vanilla completion model over a KG that combines all facts without alignment, underscoring the value of joint training for these tasks.

\end{abstract}

\section{Introduction}
\label{sec:Intro}

A knowledge graph (KG), also called as knowledge base (KB), has nodes representing entities and edges representing relations.  Entities have unique canonical IDs.  Relations also have canonical labels such as born-in or works-at, with associated IDs.  A fact triple in a KG is of the form (subject entity, relation, object entity).  A KG is usually associated with a human language.
Each entity (or relation) ID is associated with one or more surface forms in the KG. E.g., the ID for the country USA may have aliases like ``United States of America". 

KGs are usually very incomplete, as curators struggle to keep up with the real world.  KG completion (KGC) is thus a strongly motivated problem and studies the prediction of true facts unknown to the KG.  While the problem is well-researched, with dozens of approaches explored over the last few years \citep{bordes2013transe, dettmers2018conve, sun2019rotate, trouillon2016complex}, most KGC research is applicable to only one KG in one language at a time. However, different language speakers would maintain separate KGs in their own languages.  Independent completion of each KG may not be optimal, since information from one KG will likely help completion of the other. 

A second issue is that each KG will give a different ID, and often also the surface form, to the same entity, such as ``Estados Unidos de América" for the USA entity in a Spanish KG. This leads to the problem of ID proliferation. A recent line of research around entity alignment (EA) across KGs  in different languages attempts to assign a unique ID to all IDs representing the same entity \citep{Chen+2017MtransE, Sun+2017JAPE, Sun+2018BootEA, cao-etal-2019-multi, Sun+2020AliNet, chen+2021jeans, Tang+2020BertInt}. A related task is relation alignment (RA), though relatively less attention has been given to this for multilingual KGs\todo{references}. We note that RA involves \emph{global} evidence, because the decision to merge two relations in two languages can have far-reaching consequences to many facts in both KGs.

Our key contribution is to recognize and exploit the synergy between Multilingual KGC, EA and RA tasks. It is not surprising that such alignments will allow a KG an access to more facts and will lead to better completion. Similarly, if a fact can be predicted with high confidence in one KG that may give additional support to alignment of its constituent entities and relations.

\todo{give a motivating bidirectional synergy example}

In this paper, we present \shortname, a multi-task system that learns to optimize for KGC, EA and RA jointly.  At the heart of \shortname{} is the subject-object signature of each relation, which we represent as a bag of embeddings.  These bags are compared for equivalence and implication between relations, and trained via an end-to-end training protocol for the multiple tasks.


We evaluate \shortname{} on slices of DBPedia in five languages.  We compare it against a strong state-of-the-art single-KGC system trained over each monolingual KG Separately. Compared to this monolingual baseline, we find that \shortname{} achieves substantial accuracy boost due to other KGs that get better aligned by our system, obtaining 10-32 pt MRR improvements across languages.For EA and RA, We compare it against a multilingual baseline trained over the single KG that has union of all (unaligned) facts. Joint training yields 22 pt HITS@1 gain on EA, and 26 pt HITS@1 gain in RA for frequent relations. Our code and data sets will be made publicly available.

\section{Notation and preliminaries}
\label{sec:Prelims}

Throughout, we use knowledge graph (KG) and knowledge base (KB), and similarly KBC and KGC interchangeably.  A KG consists of entities $E$ and relations (aka relation types)~$R$.  A KG instance is a triple $(s,r,o)$ where $s,o\in E$ and $r\in R$.  These are all canonical IDs, but each ID is associated with aliases in one or more languages.

\subsection{KGC task}

For any single KG, training data is provided as $(s,r,o)$ triples.  A test instance has the form $(s,r,?)$ or $(?,r,o)$ where the system has to predict $o$ or~$s$.  Multiple correct values are possible.  The evaluation protocol usually has the system rank candidate $o$'s or $s$'s and then measures MRR or hits@$K$.

\subsection{Alignment task}

We consider a set L = ${[l_1, l_2...]}$ of languages.For simplicity of exposition, we consider two KGs called $\KG_{l}$ and $\KG_{l'}$.  Here $\KG_{l}$ represents the KG supported by the language l.  An entity in this KG is called $e_{l}$.  A relation in this KG is called $r_{l}$. For simplicity of exposition, we consider two KGs called $\KG_{l}$ and $\KG_{l'}$ where $l,l' \in L$.

Although we cast the alignment task as between KGs in two different languages, alignment between diverse KGs even in the same language (such as Wikipedia and IMDB) are considered in the same spirit.  Furthermore, it is possible for multiple KGs to take the place of $\KG_{l}$ to improve KGC in $\KG_{l'}$.

An equivalence between entities $e_{l}$ and $e_{l'}$ is specified as $e_{l} \equiv e_{l'}$, also written as the triple $(e_{l}, \equiv, e_{l'})$.
This induces a graph by adding some more edges with label `$\equiv$' to $\KG_{l} \cup \KG_{l'}$.
Similarly an equivalence between relations $r_{l}$ and $r_{l'}$ is specified as $r_{l} \equiv r_{l'}$; this is not easily represented in the graph $\KG_{l}\cup\KG_{l'}$, however.

Other relations may be possible between relation pairs, such as $r_{l} \implies r_{l'}$, which means, for all $s, o$ such that $(s,r_1,o)$ holds, so does $(s,r_2,o)$.

During training, a set of entity equivalences
$\{ (e^n_{l}, \equiv, e^n_{l'}): n=1,\ldots,N \}$ and a set of relation equivalences $\{ (r^m_{l}, \equiv, r^m_{l'}):  m=1,\ldots,M \}$ are revealed to the system.  The goal of the system is to infer additional entity and relation equivalences.  The system is usually called upon to produce a ranked list of equivalences, which is evaluated using HITS@$K$ or MRR.

To achieve KGC enhanced with alignment, the system has to infer additional triples in either KG, making best use of the revealed equivalences.  For the KG alignment goal, the system has to infer additional alignment triples between an entity or relation in $\KG_{l}$ and an entity or relation in~$\KG_{l'}$.

\section{Proposed methods}
\label{sec:Method}

\subsection{Baseline}

Many KG embedding and KBC algorithms have been proposed in the last few years.  ComplEx \citep{trouillon2016complex} with all negative instances (no sampling) gives the best predictions \citep{jain-baseline}, so we use it as our baseline KGC gadget.  ComplEx defines a triple score as 
\begin{align}
\label{eq:complex}
f(s,r,o) = \Re\left( \langle \bm{s}, \bm{r}, \bm{o}^\star \rangle \right),
\end{align}  
where $c^\star$ is complex conjugate, $\langle\cdots\rangle$ is a 3-way elementwise inner product and $\Re(\cdot)$ is the real part of a complex number.   When applied to any one KG in isolation, we call this method \textbf{\KbcOne{}}.

Perhaps the most straight-forward way to apply a KGC system is to compute $\KG_{l} \cup \KG_{l'}$, collapse node pairs specified as equivalent in $\{(e_{l},\equiv,e_{l'})\}$, and rename all equivalent $r_{l}, r_{l'}$ relation IDs to a common new ID. A KGC system can work on the resulting KG, and this method is called \textbf{\KbcUnion{}}.  Of course, this scheme does not impute any equivalences beyond what are explicitly provided.

\subsection{Joint alignment and completion}

For both textual and structured inputs, the problem of inferring a predicate or relation as entailed by another has been studied \citep{lin2001dirt, bhagat-etal-2007-ledir, nakashole-etal-2012-patty}.  With that work as our point of departure, we ask: How similar are two relations $r_1, r_2$ in one KG?

To build an estimate of similarity, we define the subject-object signature of a relation wrt a \KG:
\begin{align}
  \SO(r) &= \{ (s,o):  (s,r,o) \in \KG \}
\end{align}
Here $s,o$ are interpreted as canonical IDs.

\subsubsection{Jaccard similarity}

Jaccard similarity can then be used as a standard symmetric belief that two relations are equivalent:
\begin{align}
b(r_1\Leftrightarrow r_2) = \frac{|\SO(r_1)\cap\SO(r_2)|}{|\SO(r_1)\cup\SO(r_2)|}.
\end{align}

We add a threshold on these scores to reduce noise of false relation alignment signal.

\subsubsection{Asymmetric subsumption}

The issue with Jaccard similarity is that it can give a symmetric high score to relation pairs having asymmetric implications between them.  E.g., in the DBP5L data set, Jaccard similarity gives a large similarity score between \path{locationCity} and \path{headquarter}, or \path{keyPerson} and \path{founders}.

Therefore, we need a belief measure for one relation subsuming another, which we define as \todo{other fuzzy defns exist, discuss}
\begin{align}
b(r_1{\implies}r_2) &=
\frac{|\SO(r_1)\cap \SO(r_2)|}{|\SO(r_1)|} \in[0,1]
\label{eq:AsymEntail}
\end{align}
$b(r_2{\implies}r_1)$ is defined likewise.  Extending the logical statement
\begin{align}
(r_1 \Leftrightarrow r_2) \quad &
\text{iff}\quad
(r_1{\implies}r_2) \wedge (r_2{\implies}r_1)
\intertext{to the fuzzy domain, we get}
b(r_1\Leftrightarrow r_2) &=
\min\left\{
b(r_1{\Rightarrow}r_2), b(r_2{\Rightarrow}r_1)
\right\}  \label{eq:AsymEquiv}
\end{align}

\subsubsection{Soft subsumption and equivalence}

In the above definitions involving $\SO$, we assumed the $(s,o)$ pairs were represented using canonical entity IDs.  This may not be useful in the KG alignment task because canonical entity IDs will frequently not match.

Recall that the KBC system obtains embedding vectors $\bm{e}$ for each entity $e$ in the KG.  Reusing notation, we modify our earlier definition of subject-object signature to
\begin{align}
\SO(r) &= \{ (\bm{s}, \bm{o}):
(s,r,o) \in \KG \},  \label{eq:softSO}
\end{align}
i.e., where each element is the concatenation of the subject and object \emph{embedding vectors}. \todo{unclear if we should dissolve $s,o$ boundary right here}

Now consider one relation from each of two KGs to be aligned, viz., $r_{l}, r_{l'}$.  Suppose the $(\bm{s}, \bm{o})$ pairs of $r_{l}$ are indexed by $i$ and pairs of $r_{l'}$ are indexed by~$j$.  To generalize \eqref{eq:AsymEquiv}, we build a matrix $A_{r_{l},r_{l'}}$ of pairwise cosine similarities:
\begin{align}
A_{r_{l},r_{l'}}[i,j] &=
\cos\!\big(\!\SO(r_{l})[i], \SO(r_{l'})[j]\big)
\end{align}
The continuous extension of $\SO(r_{l})\cap\SO(r_{l'})$ involves solving a maximal bipartite matching problem using $A_{r_{l},r_{l'}}$ as edge weights.  Ideally, we should be able to backpropagate various KBC and alignment losses past the solution of the matching problem to the entity and relation embeddings.  Gumbel-Sinkhorn matrix scaling \citep{Cuturi2013sinkhorn, Mena+2018GumbelSinkhorn} can be used for this purpose, but it is computationally expensive at KG scales.

\todo{is the approx good enough? (discuss in meeting)} Here we use a computationally cheaper approximation: only if $i$ is $j$'s strongest partner and $j$ is $i$'s strongest partner, we choose edge $(i,j)$ and accrue (toward the soft version of $\SO(r_{l})\cap\SO(r_{l'})$) the score increment
\begin{align}
\sigma\big(A_{r_{l},r_{l'}}[i,j]\,w + c\big),
\end{align}
where $\sigma$ is the sigmoid nonlinearity, and $w>0, b \in \mathbb{R}$ are model parameters trained along with all embeddings.  Summarizing, we estimate
\begin{multline}
|\SO(r_{l})\cap\SO(r_{l'})| = \\
\textstyle \sum_{i,j: p(i,j)} 
\sigma\big(A_{r_{l},r_{l'}}[i,j]\,w + c\big), \label{eq:overlap}    
\end{multline}
where the partner test indicator is written as
\begin{multline}
p(i,j) = 
\llbracket \textstyle 
i=\argmax_{i'} A[i',j] \rrbracket
\\[-.5ex] 
\llbracket \textstyle 
j = \argmax_{j'} A[i,j'] \rrbracket
\end{multline}
We continue to use \eqref{eq:AsymEntail} and \eqref{eq:AsymEquiv} as defined with the modified continuous definition of intersection.

\subsubsection{Assembling a joint loss function}

We start with the ComplEx KBC loss and then augment with loss terms corresponding to entity alignment and relation alignment.  
Using $f$ in Equation \ref{eq:complex}, ComplEx defines
\begin{align}
\Pr(o|s,r) &= e^{f(s,r,o)} \Big/
\textstyle \sum_{o'} e^{f(s,r,o')}, 
\label{eq:pro} \\
\Pr(s|o,r) &= e^{f(s,r,o)} \Big/
\textstyle \sum_{s'} e^{f(s',r,o)},
\label{eq:prs}
\end{align}
and the log-likelihood \textbf{KGC loss} as
$L_\text{KGC}=$
\begin{multline}
\sum_{(s,r,o)\in\KG}
-\log\Pr(o|s,r)-\log\Pr(s|o,r).
\end{multline}
\todo{(do NOT cite baseline strikes back here)}
In our experiments, we found that retaining the full negative sets $\{s'\},\{o'\}$ is better than negative sampling, which we implemented using 1-N scoring \citep{jain-baseline,dettmers2018conve}.

We have no counterpart to the \textbf{entity alignment loss} \citep{Chen+2017MtransE, Sun+2018BootEA} because, for any pair $e_{l} \equiv e_{l'}$, we force the two entities to share the same embedding vector.  But we do introduce a novel \textbf{relation alignment loss}.  If $b(r_{l} \Leftrightarrow r_{l'})$ is large, but the relation embeddings $\bm{r}_{l}$ and $\bm{r}_{l'}$ are very dissimilar, we should assess a loss.  This naturally suggests the additional relation alignment loss term $L_\text{RA1}=$
\begin{align}
\sum_{c(r_{l},r_{l'})}
b(r_{l} \Leftrightarrow r_{l'})
\big\| \bm{r}_{l} - \bm{r}_{l'} \big\|_1.
\end{align}
\todo{play with these}
Other forms are possible, like
\begin{align}
\sum_{c(r_{l},r_{l'})}
\text{BCE}\big( b(r_{l} \Leftrightarrow r_{l'}),
\sigma(\cos(\bm{r}_{l}, \bm{r}_{l'})) \big).
\end{align}

where the corresponding relation test indicator is written as 

\begin{multline}
    c(r_{l},r_{l'}) = 
    \llbracket \textstyle 
    r_{l} \equiv \argmax_{r_{l}'} b(r_{l'} \Rightarrow r_{l}') \rrbracket
    \\[-.5ex] 
    \llbracket \textstyle 
    r_{l'} \equiv \argmax_{r_{l'}'} b(r_{l} \Rightarrow r_{l'}') \rrbracket
\end{multline}

\subsubsection{Joint loss}

We put together all the above loss components and also a L2-regularizer $L_\text{reg}$ on entity and relation embeddings, each multiplied by tuned hyperparameters~$\alpha,\beta$:
\begin{align}
L_\text{KGC} + 
\alpha L_\text{reg} +
\beta L_\text{RA1} 
\end{align}

We initialize all elements of entity and relation embeddings to $\mathcal{N}(0,0.05)$.  Entity embeddings are not sensitive to alignment initially except those involved in seed alignment, hence we initialize the $c=-90$ and $w=100$ in \eqref{eq:AsymEntail}.  Thus, only very high initial cosine similarities contribute to equivalence score computation.  As a form of curriculum, we let relation alignments get stable for few iterations and then make the equivalence scores trainable.

\section{Experiments}



\subsection{Dataset}

Standard mono-lingual KGC datasets such as FB15k, FB15k-237, WN18, WN18RR, and Yago3-10, are not directly suited to evaluation of multi-lingual KGC task.
Also, the focus on KG alignment is still relatively recent, with only a few suitable multi-lingual datasets that provide gold alignments between entity and relation pairs for training and evaluation. One such dataset is DBP5L, derived from DBPedia in five languages: English~(En), Greek~(El), Spanish~(Es), Japanese~(Ja) and French~(Fr).   Typically, about \todo{@HK: must be somewhat diverse across lang pairs; provide a small table} 40\% of entities in one language are aligned to entities in the other language. We  use the same 60-30-10 splits of the KG triples into train-dev-test folds as in \citet{chen-etal-2020-multilingual} and combine the train set of all languages for training.We also pick half of seed enitity alignments randomly for training and rest is held for evaluation. 

Because DBPedia uses a uniform relation vocabulary that is normalized across all languages, it cannot be directly used to test RA. To simulate that KBs in different languages come from different sources, we declare a unique ID for each relation in a language. This creates a testbed for the RA task.
\todo[inline]{seems wikidata also has all the qualities we need; should try to make a dataset from there too (code exists).}

\begin{table}
\centering
\resizebox{\hsize}{!}{%
\tabcolsep 1pt
\begin{tabular}{|l|r|r|r|r|r|}
\hline
\textbf{Language} & \textbf{Greek} & \textbf{Japanese} & \textbf{Spanish} & \textbf{French} & \textbf{English} \\ \hline
\textbf{\#Entity}   & 5,231  & 11,805 & 12,382 & 13,176 & 13,996 \\ \hline
\textbf{\#Relation} & 111    & 128    & 144    & 178    & 831    \\ \hline
\textbf{\#Triples}  & 13,839 & 28,774 & 54,066 & 49,015 & 80,167 \\ \hline
\end{tabular}%
}
\caption{Salient statistics of KGs.}
\label{tab:SalientStats}
\end{table}

\subsubsection{Salient statistics}

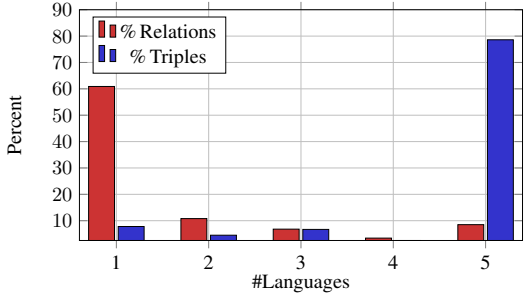
\begin{figure}[h]
\centering
\resizebox{.9\hsize}{!}{%
\begin{tikzpicture}
\begin{axis}[
legend pos=north west,
height=6cm, width=10cm,
grid=both,
    symbolic x coords={1,2,3,4,5},
    ybar,
    enlarge y limits={value=1},
    ylabel={Percent},
    xlabel={\#Languages},
    xtick = data,
    ytick = {0,10,20,30,40,50,60,70,80,90},
    ymax=90,
    bar width=14pt,
]
\addplot [ybar,fill=red!60!gray]
	coordinates {(1,60.9) (2,10.8)
		 (3,6.8) (4,3.4) (5,8.5)};
\addplot [ybar,fill=blue!60!gray]
	coordinates {(1,7.8) (2,4.5)
		 (3,6.7) (4,2.5) (5,78.6)};
\legend{\%\,Relations,\%\,Triples}
\end{axis}
\end{tikzpicture}
}
\caption{Fraction of relation labels and their fact tuples that are available in various number of languages between 1 and~5.}
\label{fig:RelDist}
\end{figure}

\tablename~\ref{tab:SalientStats} lists the statistics of the five KGs in DBP5L.  \En{} is expectedly the most well-populated.
\figurename~\ref{fig:RelDist} shows that 60\% of the relation labels have associated string descriptions in only one of five languages mostly English in DBP-5L.  However, these relation labels account for only 8\% of fact triples.  Meanwhile, almost 80\% of fact triples have relations expressed in all five languages.

\subsection{Performance measures}
 
We evaluate \shortname{} on DBP5L dataset, on three different tasks --- KGC, EA and RA. 
As is common when evaluating the system for the task of KGC, we regard test instances $(s,r,?)$ \todo{PJ: do both for next version - (?,r,o) and (s,r,?)} as a task of ranking $o$ (on the basis of scores computed in equations \eqref{eq:pro} and \eqref{eq:prs}), with gold $o^*$ known. We report MRR (Mean Reciprocal Rank) and the fraction of queries where $o^*$ is recalled within rank 1 and rank 10 (HITS). The \emph{filtered} evaluation removes valid train or test tuples ranking above $(s,r,o^*)$ for scoring purposes.

 
To evaluate the system for EA, we use the test instance  $(e_{l}, \equiv, ?)$ as a task of ranking $e_{l'}$, using the cosine distance between the entity embeddings of the language pair. We calculate Hits@1 and Hits@10 on the resulting rankings.

Similarly, to evaluate the system on the task of RA, we use the test instance  $(r_{l}, \equiv, ?)$ as a task of ranking $r_{l'}$, using the cosine distance between the relation embeddings of the language pair. We calculate Hits@1 and Hits@10 on the resulting rankings.

Note that we use l and l' as placeholders for various language pairs we train/test on.

\subsection{KGC performance}

\tablename~\ref{tab:Learnable Asymmetric Performance} shows the KGC performance on the five languages of DBP-5L namely Greek, Japanese, French, Spanish and English.
Though train and test set of each language KG have no overlap ($\KG_{l}(train) \cap \KG_{l}(test) = \varnothing$). 
But because we assign the same IDs for aligned entities during training, some form of overlap may exist in the combined training triples of all languages with test sets in any language.  E.g.,  if $(s_{l},r_{l},o_{l}) \in \KG_{l}(train)$ and $(s_{l'},r_{l'},o_{l'}) \in \KG_{l'}(test)$, where $s_{l} \equiv s_{l'}$, $o_{l} \equiv o_{l'}$ and $r_{l} \equiv r_{l'}$ (relation alignment may or may not be used).

Therefore, rather than report metrics only on whole test-split, which can be misleading, we split our test set into two components: \textbf{seen} test set with triples/facts as shown above, and \textbf{unseen} test as the remaining triples. Qualitatively, performance on the seen split represents the capacity of the model to memorize known facts in a language and align them to another language, whereas the unseen split gives a true sense of a system's inference capability, since the fact has not been read in \emph{any} language at train time.

\tablename~\ref{tab:Learnable Asymmetric Performance} shows the results of various \shortname{} variations compared to the baseline. \KbcUnion{} being our multilingual baseline shows gains of combining the different language KGs over the monolingual baseline \KbcOne{}. We find that each modification (Jaccard, Asymmetric scores and Soft Asymmetric scores) improves the results successively. Further analysis reveals that Jaccard signficantly helps with seen split, since it is able to learn similar embeddings for two aligned relations and entities, and yields high scores for seen tuple in a different language. Asymmetric computations help further as they remove false positives in RA that appear in the Symmetric Jaccard version. 

  
Finally, making these asymmetric implication scores trainable learns better implications scores, which significantly improve entity alignment, which in turn gives better relation alignment. So we get best results on the both unseen and seen test splits for this version.

\begin{table*}
\centering
\resizebox{\hsize}{!}{%
        \begin{tabular}{|
        c |
        c |
        c |
        c |
        c |
        c |
        c |
        c |
        c |
        c |
        c |
        c |
        c |
        c |
        c |
        c |}
        \hline
         &
          \multicolumn{3}{c|}{\textbf{GREEK}} &
          \multicolumn{3}{c|}{\textbf{JAPANESE}} &
          \multicolumn{3}{c|}{\textbf{FRENCH}} &
          \multicolumn{3}{c|}{\textbf{SPANISH}} &
          \multicolumn{3}{c|}{\textbf{ENGLISH}} \\ \hline
         &
          \textbf{Hits@1} &
          \textbf{Hits@10} &
          \textbf{MRR} &
          \textbf{Hits@1} &
          \textbf{Hits@10} &
          \textbf{MRR} &
          \textbf{Hits@1} &
          \textbf{Hits@10} &
          \textbf{MRR} &
          \textbf{Hits@1} &
          \textbf{Hits@10} &
          \textbf{MRR} &
          \textbf{Hits@1} &
          \textbf{Hits@10} &
          \textbf{MRR} \\ \hline
          \textbf{\KbcOne{}} &
          23.6 &
          49.0 &
          31.9 &
          26.4 &
          49.4 &
          34.4 &
          24.0 &
          50.4 &
          32.8f &
          22.1 &
          50.1 &
          31.4 &
          18.8 &
          43.0 &
          26.9 \\ \hline
        \textbf{\KbcUnion{}} &
          39.0 &
          75.9 &
          52.1 &
          40.3 &
          69.9 &
          50.9 &
          37.5 &
          70.3 &
          48.5 &
          34.9 &
          66.3 &
          45.6 &
          25.3 &
          51.7 &
          34.3 \\ \hline
        \textbf{\shortname(Jaccard)} &
          49.9 &
          80.8 &
          61.1 &
          46.6 &
          73.6 &
          56.1 &
          41.5 &
          72.6 &
          52.3 &
          40.4 &
          69.5 &
          50.4 &
          27.1 &
          53.9 &
          36.4 \\ \hline
        \textbf{\shortname(Asymmetric)} &
          51.1 &
          \textbf{83.8} &
          62.9 &
          47.6 &
          74.4 &
          56.8 &
          41.8 &
          72.4 &
          52.5 &
          40.2 &
          69.2 &
          50.0 &
          26.5 &
          53.5 &
          35.6 \\ \hline
        \textbf{\shortname(Soft Asymmetric)} &
          \textbf{53.4} &
          83.6 &
          \textbf{64.6} &
          \textbf{49.81} &
          \textbf{75.9} &
          \textbf{58.9} &
          \textbf{44.1} &
          \textbf{73.9} &
          \textbf{54.8} &
          \textbf{43.3} &
          \textbf{70.7} &
          \textbf{52.8} &
          \textbf{28.6} &
          \textbf{55.4} &
          \textbf{37.5} \\ \hline
        \multicolumn{16}{|c|}{\textbf{UNSEEN TEST SET}} \\ \hline
        \textbf{\KbcUnion{}} &
          25.4 &
          64.3 &
          38.5 &
          27.0 &
          57.7 &
          37.7 &
          27.3 &
          61.1 &
          38.25 &
          22.7 &
          55.2 &
          33.5 &
          19.4 &
          45.6 &
          28.1 \\ \hline
        \textbf{\shortname(Jaccard)} &
          28.6 &
          67.3 &
          42.3 &
          30.9 &
          61.4 &
          41.0 &
          28.7 &
          62.8 &
          40.1 &
          25.7 &
          58.5 &
          36.6 &
          19.9 &
          47.6 &
          29.4 \\ \hline
        \textbf{\shortname(Asymmetric)} &
          28.6 &
          \textbf{71.9} &
          43.1 &
          29.8 &
          62.1 &
          40.4 &
          27.9 &
          62.5 &
          39.5 &
          25.1 &
          58.2 &
          35.8 &
          18.9 &
          47.0 &
          28.1 \\ \hline
        \textbf{\shortname(Soft Asymmetric)} &
          \textbf{29.9} &
          71.2 &
          \textbf{44.3} &
          \textbf{32.8} &
          \textbf{64.2} &
          \textbf{43.1} &
          \textbf{29.2} &
          \textbf{64.3} &
          \textbf{41.3} &
          \textbf{26.9} &
          \textbf{59.6} &
          \textbf{37.8} &
          \textbf{20.5} &
          \textbf{48.9} &
          \textbf{29.8} \\ \hline
        \multicolumn{16}{|c|}{\textbf{SEEN TEST SET}} \\ \hline
        \textbf{\KbcUnion{}} &
          56.1 &
          90.5 &
          69.2 &
          66.3 &
          93.7 &
          76.8 &
          64.3 &
          94.4 &
          75.3 &
          66.6 &
          95.3 &
          77.2 &
          65.8 &
          93.5 &
          76.3 \\ \hline
        \textbf{\shortname(Jaccard)} &
          76.5 &
          97.8 &
          84.8 &
          77.3 &
          97.4 &
          85.6 &
          74.9 &
          98.3 &
          84.4 &
          78.8 &
          98.3 &
          86.5 &
          76.2 &
          97.9 &
          84.7 \\ \hline
        \textbf{\shortname(Asymmetric)} &
          79.6 &
          98.7 &
          87.7 &
          82.5 &
          98.5 &
          89.0 &
          78.3 &
          98.3 &
          86.4 &
          79.7 &
          98.2 &
          87.0 &
          78.5 &
          98.3 &
          86.5 \\ \hline
        \textbf{\shortname(Soft Asymmetric)} &
          \textbf{82.9} &
          \textbf{99.1} &
          \textbf{89.9} &
          \textbf{83.2} &
          \textbf{98.6} &
          \textbf{89.8} &
          \textbf{83.1} &
          \textbf{99.1} &
          \textbf{90.0} &
          \textbf{86.0} &
          \textbf{99.7} &
          \textbf{91.8} &
          \textbf{83.3} &
          \textbf{99.4} &
          \textbf{90.3} \\ \hline
        \end{tabular}%
        }
    \caption{KGC Performance of Models on Five languages}
    \label{tab:Learnable Asymmetric Performance}
\end{table*}

\subsection{Alignment performance}

\tablename~\ref{tab:Alignment Performance} reports the performance of the various models on RA and EA tasks.We use entity and relation alignment results of \KbcUnion{} as our baseline .We find that all \shortname{} variants outperform the baseline by vast margins. We notice that performance gain in RA are higher for highly frequent relations, and for less frequent relations, decreases slightly compared to Jaccard. Since a much larger fraction of triples (see \figurename~\ref{fig:RelDist}) express these highly frequent relations, it overall improves the performance of KGC significantly. 


We also note that Jaccard and Asymmetric Jaccard models perform Entity alignment based on KGC loss only, since they do not have any trainable alignment loss. Since those models significantly improve the performance of EA, it provides evidence that KGC can result in better alignment. Learnable asymmetric model, of course, incentivizes better alignment too, leading to a further improvement in performance.

    
\begin{table*}
\centering
\adjustbox{max width=.8\hsize}{%
\begin{tabular}{|c|c|c|c|c|c|c|}
\hline
\multicolumn{1}{|l|}{} &
  \multicolumn{2}{c|}{\textbf{Relation Alignment(\textless{}500)}} &
  \multicolumn{2}{c|}{\textbf{Relation Alignment(${\ge}500$)}} &
  \multicolumn{2}{c|}{\textbf{Entity Alignment}} \\ \hline
\multicolumn{1}{|l|}{}       & \textbf{Hits@1} & \textbf{Hits@3} & \textbf{Hits@1} & \textbf{Hits@3} & \textbf{Hits@1} & \textbf{Hits@10} \\ \hline
\textbf{\KbcUnion{}}            & 19.8            & 28.7            & 42.8            & 66.5            & 23.5            & 42.8             \\ \hline
\textbf{\shortname(Jaccard)}  & \textbf{27.8}   & \textbf{39.8}   & 53.5            & 72.7            & 38.3            & 55.8             \\ \hline
\textbf{\shortname(Asymmetric)}    & 27.2            & 38.7            & 66.9            & 75.0            & 40.7            & 57.4             \\ \hline
\textbf{\shortname(Soft Asymetric)} & 26.6            & 37.6            & \textbf{68.8}   & \textbf{76.5}   & \textbf{45.3}   & \textbf{61.9}    \\ \hline
\end{tabular}%
}
\caption{Entity and relation alignment Performance of Models on Five languages}
\label{tab:Alignment Performance}
\end{table*}

\section{Related work}
\label{sec:Rel}

\subsection{KBC}

KBC through learning embeddings for KG artifacts is a \href{https://paperswithcode.com/task/knowledge-graph-completion}{densely-populated research landscape}.  Among the best performers are ComplEx \citep{jain-baseline, trouillon2016complex}, ConvE \cite{dettmers2018conve}, and RotatE \citep{sun2019rotate}.  Almost all such systems are designed for a single KG, or are agnostic to the language used in entity and relation aliases.

\subsection{KG alignment}

Although more recent, interest in \href{https://paperswithcode.com/task/entity-alignment}{KG alignment} is rapidly growing.  MTransE \citep{Chen+2017MtransE}, as the name suggests, uses TransE \citep{bordes2013transe} on each KG separately, and adds a loss term that penalizes large distance between embeddings of equivalent entities.
BootEA \citep{Sun+2018BootEA} finds entity embeddings in their respective KGs and estimates a probability of equivalence by comparing their embeddings.  This probability is then used for a semi-supervised bootstrapping of equivalent pairs.
JAPE \citep{Sun+2017JAPE} builds attribute- and network neighborhood-based embeddings of entities and combines them with EA constraints.
MuGNN \citep{cao-etal-2019-multi} uses a graph neural network (GNN) to embed entities informed by entailment constraints of the form $(s,r,o)\implies(s,r',o)\forall s,o$.  These constraints \todo{check} appear to be manually provided and are the only KBC mechanism.  Afterward, tied GNNs obtain node embeddings which are compared to propose $\equiv$ links.
AliNet \citep{Sun+2020AliNet} is another GNN-based EA system.  It uses an attention mechanism over larger node neighborhoods to build node representations.
MultiKE \citep{Zhang+2019MultiKE} is perhaps the only work that treats entity and relation alignments at par and combines multiple views to make decisions. However, they use relation names rather than their structural subject-object summaries.
JEANS \citep{chen+2021jeans} is distinctive in that it links (`grounds') a text corpus to enities and uses TransE \citep{bordes2013transe} for the KG, skip-grams for the text, with additional alignment constraints as usual.
BERT-INT \citep{Tang+2020BertInt} is another EA system that combines mBERT-obtained features from entity aliases and text descriptions with soft 1-hop graph neighborhood matching.


Except MultiKE, most systems focus on EA, assuming RA is unnecessary, or already accomplished. \citet{chen-etal-2020-multilingual} state so explicitly.
\todo{reviewer may ask why we did not use the best EA system in \shortname}
At a high level, EA is a component of \shortname, and the precise EA method used is orthogonal to \shortname{} itself.


\section{Conclusion}
\label{sec:End}

We have presented \shortname, a system that jointly learns to complete multiple KGs (KGC) and align their entities and relations.  KGC and entity alignment were known tasks, but relation alignment was, to our knowledge, never integrated with them.  In extensive experiments, \shortname{} significantly improves KGC accuracy, as well as alignment scores, underscoring the value of joint alignment and completion.

\bibliography{ALIGNKBC,anthology}

\begin{thebibliography}{19}
\expandafter\ifx\csname natexlab\endcsname\relax\def\natexlab#1{#1}\fi

\bibitem[{Bhagat et~al.(2007)Bhagat, Pantel, and Hovy}]{bhagat-etal-2007-ledir}
Rahul Bhagat, Patrick Pantel, and Eduard Hovy. 2007.
\newblock \href {https://www.aclweb.org/anthology/D07-1017} {{LEDIR}: An
  unsupervised algorithm for learning directionality of inference rules}.
\newblock In \emph{Proceedings of the 2007 Joint Conference on Empirical
  Methods in Natural Language Processing and Computational Natural Language
  Learning ({EMNLP}-{C}o{NLL})}, pages 161--170, Prague, Czech Republic.
  Association for Computational Linguistics.

\bibitem[{Bordes et~al.(2013)Bordes, Usunier, Garcia-Duran, Weston, and
  Yakhnenko}]{bordes2013transe}
Antoine Bordes, Nicolas Usunier, Alberto Garcia-Duran, Jason Weston, and Oksana
  Yakhnenko. 2013.
\newblock Translating embeddings for modeling multi-relational data.
\newblock In \emph{NeurIPS}, pages 1--9.

\bibitem[{Cao et~al.(2019)Cao, Liu, Li, Liu, Li, and
  Chua}]{cao-etal-2019-multi}
Yixin Cao, Zhiyuan Liu, Chengjiang Li, Zhiyuan Liu, Juanzi Li, and Tat-Seng
  Chua. 2019.
\newblock \href {https://doi.org/10.18653/v1/P19-1140} {Multi-channel graph
  neural network for entity alignment}.
\newblock In \emph{Proceedings of the 57th Annual Meeting of the Association
  for Computational Linguistics}, pages 1452--1461, Florence, Italy.
  Association for Computational Linguistics.

\bibitem[{Chen et~al.(2017)}]{Chen+2017MtransE}
Muhao Chen et~al. 2017.
\newblock \href {https://www.ijcai.org/proceedings/2017/0209.pdf} {Multilingual
  knowledge graph embeddings for cross-lingual knowledge alignment}.
\newblock In \emph{IJCAI}, page 1511–1517.

\bibitem[{Chen et~al.(2021)}]{chen+2021jeans}
Muhao Chen et~al. 2021.
\newblock \href {https://arxiv.org/pdf/2005.00171} {Cross-lingual entity
  alignment with incidental supervision}.
\newblock In \emph{EACL Conference}.

\bibitem[{Chen et~al.(2020)Chen, Chen, Fan, Uppunda, Sun, and
  Zaniolo}]{chen-etal-2020-multilingual}
Xuelu Chen, Muhao Chen, Changjun Fan, Ankith Uppunda, Yizhou Sun, and Carlo
  Zaniolo. 2020.
\newblock \href {https://doi.org/10.18653/v1/2020.findings-emnlp.290}
  {Multilingual knowledge graph completion via ensemble knowledge transfer}.
\newblock In \emph{Findings of the Association for Computational Linguistics:
  EMNLP 2020}, pages 3227--3238, Online. Association for Computational
  Linguistics.

\bibitem[{Cuturi(2013)}]{Cuturi2013sinkhorn}
Marco Cuturi. 2013.
\newblock \href
  {https://papers.nips.cc/paper/4927-sinkhorn-distances-lightspeed-computation-of-optimal-transport.pdf}
  {Sinkhorn distances: Lightspeed computation of optimal transport}.
\newblock In \emph{NeurIPS}, pages 2292--2300.

\bibitem[{Dettmers et~al.(2018)Dettmers, Minervini, Stenetorp, and
  Riedel}]{dettmers2018conve}
Tim Dettmers, Pasquale Minervini, Pontus Stenetorp, and Sebastian Riedel. 2018.
\newblock \href
  {https://ojs.aaai.org/index.php/AAAI/article/download/11573/11432}
  {Convolutional 2d knowledge graph embeddings}.
\newblock In \emph{AAAI Conference}.

\bibitem[{Jain et~al.(2020)Jain, Rathi, Mausam, and
  Chakrabarti}]{jain-baseline}
Prachi Jain, Sushant Rathi, Mausam, and Soumen Chakrabarti. 2020.
\newblock Knowledge base completion: Baseline strikes back (again).
\newblock \emph{CoRR}, abs/2005.00804.

\bibitem[{Lin and Pantel(2001)}]{lin2001dirt}
Dekang Lin and Patrick Pantel. 2001.
\newblock \href {https://dl.acm.org/doi/pdf/10.1145/502512.502559} {{DIRT}:
  Discovery of inference rules from text}.
\newblock In \emph{SIGKDD Conference}, pages 323--328.

\bibitem[{Mena et~al.(2018)Mena, Belanger, Linderman, and
  Snoek}]{Mena+2018GumbelSinkhorn}
Gonzalo Mena, David Belanger, Scott Linderman, and Jasper Snoek. 2018.
\newblock \href {https://arxiv.org/pdf/1802.08665.pdf} {Learning latent
  permutations with gumbel-sinkhorn networks}.
\newblock \emph{arXiv preprint arXiv:1802.08665}.

\bibitem[{Nakashole et~al.(2012)Nakashole, Weikum, and
  Suchanek}]{nakashole-etal-2012-patty}
Ndapandula Nakashole, Gerhard Weikum, and Fabian Suchanek. 2012.
\newblock \href {https://www.aclweb.org/anthology/D12-1104} {{PATTY}: A
  taxonomy of relational patterns with semantic types}.
\newblock In \emph{Proceedings of the 2012 Joint Conference on Empirical
  Methods in Natural Language Processing and Computational Natural Language
  Learning}, pages 1135--1145, Jeju Island, Korea. Association for
  Computational Linguistics.

\bibitem[{Sun et~al.(2017)Sun, Hu, and Li}]{Sun+2017JAPE}
Zequn Sun, Wei Hu, and Chengkai Li. 2017.
\newblock \href {https://arxiv.org/abs/1708.05045} {Cross-lingual entity
  alignment via joint attribute-preserving embedding}.
\newblock In \emph{ISWC}, pages 628--644.

\bibitem[{Sun et~al.(2018)}]{Sun+2018BootEA}
Zequn Sun et~al. 2018.
\newblock \href {https://www.ijcai.org/Proceedings/2018/0611.pdf}
  {Bootstrapping entity alignment with knowledge graph embedding}.
\newblock In \emph{IJCAI}, volume~18, pages 4396--4402.

\bibitem[{Sun et~al.(2020)}]{Sun+2020AliNet}
Zequn Sun et~al. 2020.
\newblock \href {https://doi.org/10.1609/aaai.v34i01.5354} {Knowledge graph
  alignment network with gated multi-hop neighborhood aggregation}.
\newblock \emph{AAAI}, 34:222--229.

\bibitem[{Sun et~al.(2019)Sun, Deng, Nie, and Tang}]{sun2019rotate}
Zhiqing Sun, Zhi-Hong Deng, Jian-Yun Nie, and Jian Tang. 2019.
\newblock \href {https://arxiv.org/pdf/1902.10197} {Rotate: Knowledge graph
  embedding by relational rotation in complex space}.
\newblock \emph{arXiv preprint arXiv:1902.10197}.

\bibitem[{Tang et~al.(2020)}]{Tang+2020BertInt}
Xiaobin Tang et~al. 2020.
\newblock \href {https://doi.org/10.24963/ijcai.2020/439} {{BERT-INT}: A
  {BERT}-based interaction model for knowledge graph alignment}.
\newblock In \emph{IJCAI}, pages 3174--3180.

\bibitem[{Trouillon et~al.(2016)Trouillon, Welbl, Riedel, Gaussier, and
  Bouchard}]{trouillon2016complex}
Th{\'e}o Trouillon, Johannes Welbl, Sebastian Riedel, {\'E}ric Gaussier, and
  Guillaume Bouchard. 2016.
\newblock \href {http://proceedings.mlr.press/v48/trouillon16.pdf} {Complex
  embeddings for simple link prediction}.
\newblock In \emph{ICML}, pages 2071--2080.

\bibitem[{Zhang et~al.(2019)}]{Zhang+2019MultiKE}
Qingheng Zhang et~al. 2019.
\newblock \href {https://doi.org/10.24963/ijcai.2019/754} {Multi-view knowledge
  graph embedding for entity alignment}.
\newblock In \emph{IJCAI}, pages 5429--5435.

\end{thebibliography}
\bibliographystyle{acl_natbib}

\end{document}